\documentclass{article}
\usepackage{spconf,amsmath,graphicx,multirow,enumitem, url}
\usepackage{tipa}

\title{A TRANSFER LEARNING APPROACH FOR PRONUNCIATION SCORING}
\name{Marcelo Sancinetti$^1$, Jazm\'{i}n Vidal$^{1,2}$, Cyntia Bonomi$^1$, Luciana Ferrer$^2$\vspace{-0.1cm}\thanks{
This work has been funded by Argentina's National Agency for Scientific and Technological Promotion (ANPCYT) PICT 2019-01936 and by the European Union’s Horizon 2020 research and innovation program under the Marie Skłodowska-Curie grant agreement No 101007666, the Agency is not responsible for this results or use that may be made of the information.}}
\address{$^1$Departamento de Computaci\'{o}n, FCEyN, Universidad de Buenos Aires (UBA), Argentina\\
$^2$Instituto de Investigaci\'{o}n en Ciencias de la Computaci\'{o}n (ICC), CONICET-UBA, Argentina}
\begin{document}
%
\maketitle
\setlength{\abovedisplayskip}{6pt}
\setlength{\belowdisplayskip}{6pt}

\begin{abstract}
\vspace{-0.1cm}
Phone-level pronunciation scoring is a  challenging task, with performance far from that of human annotators. Standard systems generate a score for each phone in a phrase using models trained for automatic speech recognition (ASR) with native data only. Better performance has been shown when using systems that are trained specifically for the task using  non-native data. Yet, such systems face the challenge that datasets labelled for this task are scarce and usually small. In this paper, we present a transfer learning-based approach that leverages a model trained for ASR, adapting it for the task of pronunciation scoring. We analyze the effect of several  design choices and compare the performance with a state-of-the-art goodness of pronunciation (GOP) system. Our final system is 20\% better than the GOP system on EpaDB, a database for pronunciation scoring research, for a cost function that prioritizes low rates of unnecessary corrections.
\end{abstract}
%
\vspace{-0.2cm}
\section{Introduction}
\label{sec:intro}
\vspace{-0.3cm}
Computer aided pronunciation training (CAPT) systems give students feedback about the quality of their pronunciation~\cite{franco1997automatic, wei2009new} and have been shown to have a positive impact on their learning and motivation \cite{tejedor2020}. 

CAPT methods for phone-level pronunciation scoring can be classified in two groups depending on whether or not non-native data 
with pronunciation labels is used for training the system. Those that do not require such data typically rely on Automatic Speech Recognition (ASR) systems trained with native speech only. Pronunciation scores are generated using different measures of the similarity between the student's speech and native-sounding speech, represented by the ASR model \cite{franco1997automatic, kim1997automatic, witt2000phone}. 
Systems that use non-native data can be directly trained to distinguish correctly- from incorrectly-pronounced segments and use a variety of input features and classifiers \cite{wei2009new, franco2014adaptive}. They usually perform better than those of the first group \cite{chen2016computer}, but have the disadvantage of requiring non-native datasets labelled with pronunciation quality at phone level. Such datasets are usually small, making the task of training these systems challenging.

Recently, progress in deep learning for ASR \cite{hinton2012deep} triggered new work on applying deep neural networks (DNNs) for pronunciation scoring \cite{hu2015improved, huang2017transfer, duantransfer:2017, shi2020context, sudhakara2019improved}, obtaining improvements over traditional methods for both the above-mentioned groups. Notably, methods of the second group usually rely on transfer learning approaches to mitigate the problem of data scarcity.
In \cite{hu2015improved}, authors proposed to leverage an ASR model trained on native data as a feature extractor for a downstream model trained to classify mispronounced phones. Later, \cite{huang2017transfer} proposed an approach where the output layer of the ASR DNN was replaced with a new layer trained to detect incorrectly pronounced phones. Similarly, \cite{nazir2019} started from a CNN trained with a large dataset of images and trained it to classify mispronounced phones from spectrograms. 

In this work, we explore the use of a simple transfer learning-based approach for pronunciation scoring. As in \cite{huang2017transfer}, we replace the last layer of an ASR DNN, and train the resulting model for the pronunciation scoring task. 
We show that large gains can be achieved with this procedure compared to the standard GOP approach after careful selection of design choices, including the fine-tuning of an additional layer in the original DNN and the use of a loss function that compensates for the inherent imbalance across phones and classes present in  pronunciation scoring datasets.

We measure performance using the area under the curve (AUC). In addition, we propose to use an alternative cost function designed to encourage low false correction rates, something the community agrees to be essential for the practical use of these systems \cite{neri2002pedagogy}. This cost allows us to analyze the impact of the choice of decision threshold; something of utmost importance since a bad choice of threshold can result in very poor performance.
Our results on  EpaDB \cite{vidal2019epadb,vidal2021} show that the fine-tuned system achieves gains over the GOP system of 20\% on this cost. 
The main contributions of this paper are: the analysis of the best configuration to fine-tune an ASR-based pronunciation scoring system following the method proposed in \cite{huang2017transfer}, the use of a cost function as calibration-sensitive evaluation metric, and the use of our publicly available dataset. The code to obtain the results and a link to EpaDB can be found at \url{https://github.com/MarceloSancinetti/epa-gop-pykaldi}.

\section{Methods}
\label{sec:methods}
\vspace{-0.2cm}
In this section we describe the Goodness of Pronunciation (GOP) algorithm used as the baseline system and the transfer learning-based approach studied in this work.

\vspace{-0.25cm}
\subsection{DNN-based GOP system}
\vspace{-0.15cm}
The Goodness of Pronunciation (GOP) method \cite{witt2000phone} estimates scores for each phone in a phrase as the posterior probabilities of the target phones (i.e., the phones the student should pronounce) computed using the acoustic model from an ASR system trained only on native data.
Traditionally, GOP scores were computed using GMM-based acoustic models. In recent years, though, a series of papers have shown significant improvements from the use of DNN-based acoustic models \cite{hu2015improved, huang2017transfer, duantransfer:2017, shi2020context, sudhakara2019improved}. In these cases, the GOP score for a target phone $p$ that starts at frame $T$ and has a length of $D$ frames is computed as
\vspace{-0.1cm}
\begin{equation}
GOP(p)=-\frac{1}{D}\sum_{t=T}^{T+D-1}\log P_t(p|O)
\label{eq:gop}
\end{equation}
where $O$ is the full sequence of features for the waveform and $P_t(p|O)$ is an estimate of the posterior probability for phone $p$ at frame $t$. The start and end frames for each target phone are obtained using a forced-aligner given the word transcription. 
Since DNNs for ASR are usually trained to produce posterior probabilities for a set of senones  \cite{subphonetic:1992} rather than phones, to get the posterior for a certain target phone given the senone posteriors from the DNN, we sum the posteriors for all senones that correspond to the target phone, as in \cite{hu2015improved}. 

\vspace{-0.2cm}
\subsection{Proposed approach}
\label{sec:prop}
\begin{figure}[h]
\begin{centering}
\hspace{-0.4cm}\includegraphics[width=0.5\textwidth]{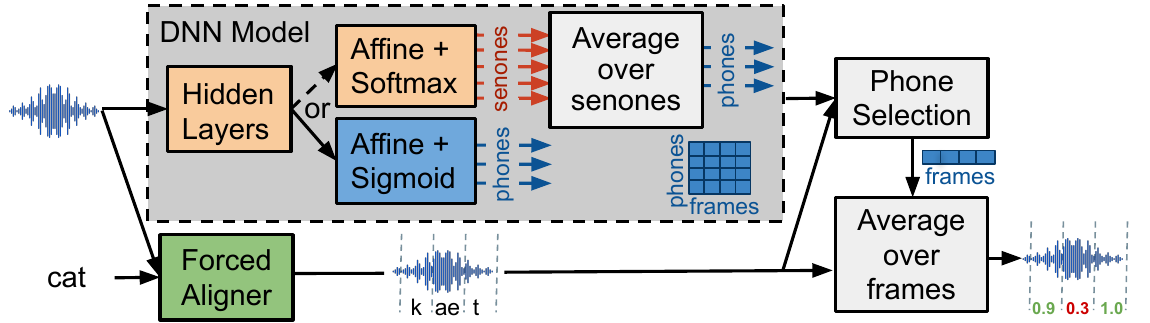}
\vspace{-0.2cm}
\caption{Schematic for the GOP and proposed models. A waveform and its transcription are fed to a forced aligner to obtain the start and end times of each target phone. A DNN model is then used to generate frame-level scores for each phone in the language. For the baseline GOP system (top branch in grey block), this block includes the step where senone posteriors are summed over all senones corresponding to each phone to get scores per phone for each frame. In the case of the proposed model (bottom branch in grey block), the DNN directly produces scores (probabilities of correct pronunciation) per phone. In the next step, the score corresponding to the phone found by the forced-aligner at each frame is selected, resulting in one score for each frame in the signal. Finally, pronunciation scores are computed by averaging the frame-level scores over all the frames for each phone in the alignments. 
}
  \label{fig:fig_method}
  \end{centering}
\end{figure}

\vspace{-0.1cm}
The GOP approach relies on a DNN that was trained for the task of senone classification, which is related to but not exactly our task of interest. Hence, in this work, we use a simple method for fine-tuning this model to the task of detecting pronunciation errors. The output layer of the senone DNN is replaced with a new layer designed to directly predict the probability of a phone being correctly pronounced. The new output layer is composed of an affine transformation followed by sigmoid activations, with one output node for each phone in the target language. The DNN with the replaced output layer is fine-tuned to optimize a binary cross-entropy loss. 
Figure \ref{fig:fig_method} shows the architecture of the proposed model, which coincides with that of the GOP baseline when the DNN model is simply the original ASR DNN followed by averaging of senone posteriors for each target phone. 

The loss function used for fine-tuning is given by:
\begin{equation}
L = - \sum_{p \in P} \sum_{y \in Y} w_{py} \sum_{t \in T_{py}} y_t \log \hat y_t + (1-y_t) \log (1-\hat y_t)
\label{eq:loss}
\end{equation}
where the first sum goes over $P$, the set of all phones in the model; the second sum goes over the two classes, $Y = \{0, 1\}$, incorrectly and correctly pronounced, respectively; the third sum goes over $T_{py} = \{t | p_t=p \land y_t=y\}$, all the frames in the waveform for which the target phone found by the forced aligner for time $t$, $p_t$, is $p$ and the pronunciation label for that frame (inherited from the label for $p_t$), $y_t$, is $y$; and $\hat y_t = P_t(p_t|O)$ is the posterior generated by the DNN for frame $t$ and phone $p_t$ given the sequence of observations $O$. The weights $w_{py}$ are used to adjust the influence of the samples from each phone and class. We evaluated two approaches: {\bf flat weights}, where all $w_{py}$ are set to 1, resulting in all frames having the same influence on the loss, and {\bf balanced weights}, where $w_{py} = 1/N_{py}$, where $N_{py}$ is the number of frames for phone $p$ and class $y$ (when $N_{py}=0$, $w_{py}$ is set to 0). The second approach is meant to compensate for the imbalance in pronunciation scoring datasets, where some phones are more frequent than others and, for most phones, one of the classes is more frequent than the other. 

We train the model using Adam optimization \cite{adam:2014}. The training loss is given by Equation (\ref{eq:loss}) averaged over all samples in a mini-batch. In the case of balanced weights, the $N_{py}$ are computed over the complete mini-batch rather than independently for each sample, to give more stability to the loss since each individual sample contains only a subset of all phones making $N_{py}= 0$ for most phones and classes on most individual samples.

 
%
In the second case, we train the model in two stages: first, only the output layer is trained over several epochs and then the second to last layer is unfrozen and both layers are further fine-tuned. In our preliminary experiments, this procedure gave better results than training both layers together from the first epoch, in agreement with many other works that do fine-tuning on different tasks \cite{howard2018universal}.
Training three layers instead of two did not lead to further improvements, probably due to the third-to-last layer having too many parameters for the amount of training data available.

We also explore different approaches that have been shown in other works and tasks to be effective during training. First, we use dropout even in the layers that have been frozen. This can be considered as a sort of data augmentation since the activations that reach the layers that we do train will not be the same each time the same sample is seen throughout the epochs, effectively creating new samples from each original one. We also explore the introduction of batch normalization in the new output layer and compare exponentially decaying and fixed learning rates.

\vspace{-0.2cm}
\section{Experimental setup}
\label{sec:exp}
\vspace{-0.2cm}
We train and evaluate the methods described above on EpaDB \cite{vidal2019epadb, vidal2021}, a dataset designed for pronunciation scoring research and development consisting of 3200 English utterances produced by 50 Spanish speakers from Argentina annotated at a detailed phonetic level. 
For each waveform, correctly- and incorrectly-pronounced labels are assigned to each of the target phones determined by the forced-alignment system by aligning that sequence of target phones with the sequence of phones in the reference transcription to which the manual annotations were aligned. 

We split the 50 speakers from EpaDB in two sets: 30 speakers are used for training and development and the remaining 20 are used for final evaluation of the systems. During development we use 6-fold cross-validation, splitting the 30 development speakers in groups of 5, using 5 groups to train the model and the other group to generate scores, and rotating the test group to generate scores for all 30 speakers. The scores generated this way are then pooled and used to compute performance metrics. For the final model evaluation, we use all 30 development speakers to train a new model, which is then tested on the 20 evaluation speakers. 

For computing GOP, we recreate the official Kaldi \cite{povey2011kaldi} recipe in PyKaldi \cite{pykaldi}. We use the Kaldi Librispeech ASR model, a TDNN-F \cite{povey2018semi} acoustic model trained with 960 hours of native English speech from the LibriSpeech~\cite{panayotov2015librispeech} collection. The acoustic features are 40-dimensional Mel frequency cepstral coefficients (MFCCs). I-vectors~\cite{dehak2010front} are used to represent the speaker's characteristics. The forced-alignments for the GOP computation are created using the same TDNN-F model. This model consists of 18 hidden layers with ReLU activations, skip connections, and batch normalization in all layers but the last two. The transformation in layers 2 through 17 is factorized into a linear plus an affine operation for improved performance. Time delay is used in the internal layers for contextualization. The first 17 layers have an output dimension of 1536. The last hidden layer is linear and has output dimension of 256. 
The original output layer has a dimension of 6024, the number of senones in the Kaldi ASR model, and uses softmax activation. When fine-tuning this model, we replace this last layer with one with 39 nodes, one for each target phone, with sigmoid activation. 

Some phones in EpaDB have very few instances with incorrect pronunciation. Performance on these phones cannot be robustly estimated. Hence, during evaluation, we discard all phones with less than 50 instances of the minority class. Further, during fine-tuning, we set the weights for these phones to 0 so that they are never trained, since they could add noise to the loss, especially for the balanced case. 

We train the model using Adam optimization, with mini-batches of 32 samples over 600 epochs using a learning rate decaying every 10 epochs by a factor of 0.9, starting from 0.01. This schedule proved to be better than using a well-tuned fixed learning rate.

Both the baseline and the proposed systems generate scores for each target phone which are expected to have higher values for correctly pronounced phones than for incorrectly pronounced ones. Hard decisions can be made by comparing these scores with a threshold. 
Each possible threshold would result in a certain false positive rate (FPR) and false negative rate (FNR). 
In our results, we report the area under the false negative versus false positive rate curve (AUC). The AUC integrates the performance over all possible operating points given by different thresholds and is a very standard metric used for this and many other tasks.

In addition, we report another metric that we consider to be more appropriate for the task of pronunciation scoring, where false negatives should be minimized to avoid frustrating the student with unnecessary corrections. We define a cost given by 0.5 FPR + FNR, where FPR is penalized less than the FNR, prioritizing low FNR over low FPR. This type of cost function is widely used in speaker verification and language detection tasks \cite{van2007introduction}, where the weights are determined depending on the application scenario.
 
To compute the cost, a decision threshold is needed. One possible approach is to choose the threshold that minimizes the cost for each phone on the test data itself, resulting on the best possible cost on that data (MinCost). Selecting the optimal threshold on the test data, though, leads to optimistic estimates of the cost. Hence, for the evaluation data we also compute the cost obtained when the threshold is selected as the one that optimizes the cost on the development data for each phone. We call this the Actual Cost (ActCost).

\vspace{-0.3cm}
\section{Results}
\label{sec:results}
\vspace{-0.2cm}

\begin{figure*}[t]
\centering
\includegraphics[width=1.01\textwidth]{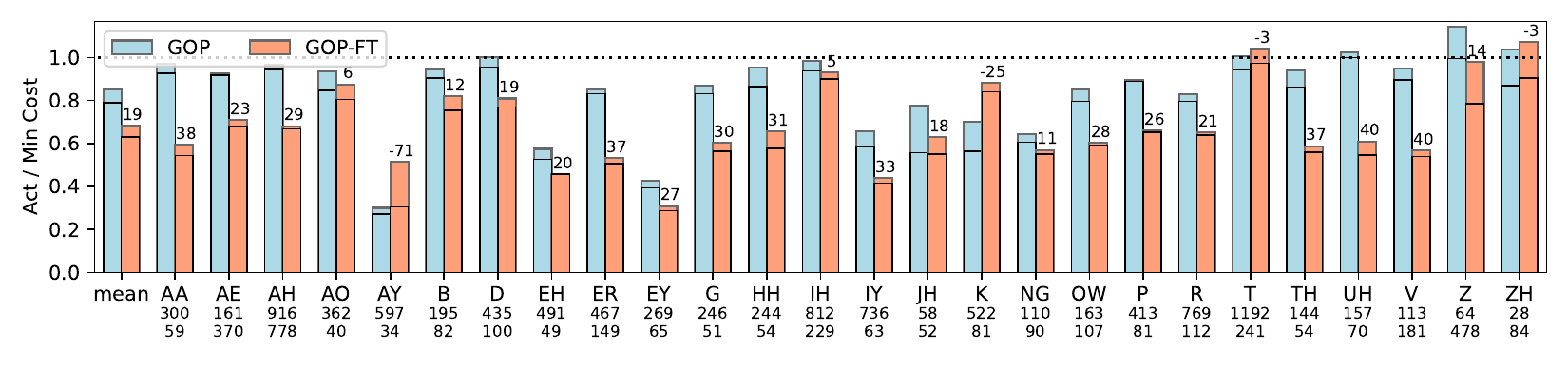}
\vspace{-1.1cm}
\caption{ActCost (bar height) and MinCost (black lines) per phone on the evaluation data. Values under the x-axis are the number of correctly and incorrectly pronounced instances of each phone. Numbers on top of the GOP-FT bars are the relative gains. 
}
\label{fig:fig_eval}
\end{figure*}

\begin{figure}[t]
\centering
\includegraphics[width=\columnwidth]{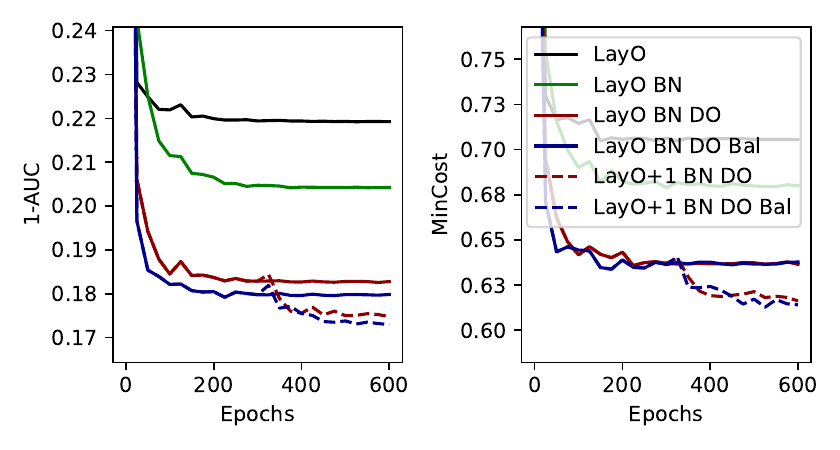}
\vspace{-1.0cm}
\caption{Average 1-AUC and MinCost over phones with more than 50 samples of each class for the development data across epochs for various fine-tuning approaches. LayO and LayO+1: the last layer or the last two layers are fine-tuned. BN: batch-normalization is used as the first component in the output layer. DO: dropout is used in all layers. Bal: the loss with balanced weights is used in training. For reference, the GOP system has 1-AUC of 0.286 and MinCost of 0.801.}
\label{fig:fig_xval}
\end{figure}

Figure \ref{fig:fig_xval} shows average MinCost and 1-AUC (so that lower values are better for both metrics)  across epochs for different fine-tuning approaches.
The first system (LayO) corresponds to fine-tuning only the output layer, without batch normalization or dropout and using the flat loss function. Adding batch normalization as the first block in the output layer gives a relative gain of 6\% in 1-AUC and 4\% in MinCost, while doing dropout during training with a probability of 0.4 gives gains of 10\% and 6\%. Finally, LayO+1, the system where the last two layers are fine-tuned, gives an additional gain of 5\% and 3\%. Using the balanced loss function gives a small but consistent gain over using the flat loss. Interestingly, the trends on 1-AUC and MinCost are similar. The best system, ``LayO+1 BN DO Bal'', gives a gain of 40\% in 1-AUC and 23\% in MinCost over the baseline GOP system on this data.

Figure \ref{fig:fig_eval} shows the results on the 20 evaluation speakers for the GOP baseline and the best fine-tuned model, which we call GOP-FT for short. 
We only report one metric due to lack of space, choosing to show the cost because it allows us to show the effect of the threshold selection. The bars with a solid black line show the MinCost, where the threshold for each phone is given by the one that optimizes the cost on the evaluation data itself. These are optimistic estimates of the cost, since, in practice, one never has the full evaluation data to estimate the thresholds on.
The top bars, on the other hand, are the ActCost, where the thresholds are estimated using the cross-validation scores, as would be done in practice.

We can see that the ActCost is within 10\% of the MinCost for most phones, indicating that the thresholds chosen on development speakers generalize well to the unseen speakers. The average FNR rate corresponding to these thresholds is 10\% and 13\%, for the GOP and GOP-FT systems, respectively, an acceptable level for real use scenarios \cite{baur2017overview, kanters2009goodness}. The average FPR is 64\% for GOP and 41\% for GOP-FT, showing a large relative improvement from the fine-tuning approach where 23\% more of the incorrect pronunciations are detected as such.  

Figure \ref{fig:fig_eval} shows a wide range of cost values across phones. In most cases, the proposed fine-tuning approach leads to gains over the GOP baseline. We hypothesize these are cases where the original ASR DNN was too permissive, allowing wrong pronunciations to get large senone posteriors for the target phone. This is corrected by fine-tuning since the system learns to distinguish what annotators considered good and bad pronunciations. 
For a few phones, the cost degrades with fine-tuning, though the degradation is relatively small in most cases. These might be cases where the model has overfitted to the training data.
Finally, note that for those phones were the cost is close to or above 1.0, the value for a naive system that always decides correct pronunciation, the system should probably not be used in practice since it would not provide useful information. Clearly, despite the gains obtained with the proposed approach, there is still work to do to improve performance on this task.

\vspace{-0.5cm}
\section{Conclusions}
\vspace{-0.3cm}

In this paper we study a simple transfer-learning approach for pronunciation scoring, where the DNN acoustic model from an ASR system is fine-tuned for the task of detecting correctly versus incorrectly pronounced phones.  
We explore several approaches for fine-tuning, showing large gains from the use of batch normalization, dropout and from the unfreezing of the last hidden layer in the DNN. 
Finally, we propose the use of a cost function designed to encourage low false correction rates, something the community agrees to be essential for the practical use of systems intended for education. 
We show that our best fine-tuned model is, on average, 20\% better in terms of cost compared to a state-of-the-art GOP system that uses the same acoustic model. 

\label{sec:refs}
\small
\bibliographystyle{IEEEbib}
\bibliography{strings,refs}

\end{document}